\documentclass[11pt]{article}

\usepackage[a4paper,margin=1in]{geometry}
\usepackage[T1]{fontenc}
\usepackage[utf8]{inputenc}
\usepackage{lmodern}
\usepackage{microtype}
\usepackage{amsmath,amssymb}
\usepackage{booktabs}
\usepackage{tabularx}
\usepackage{enumitem}
\usepackage{xcolor}
\usepackage{tikz}
\usetikzlibrary{arrows.meta,positioning,calc}
\usepackage[numbers,sort&compress]{natbib}
\usepackage[colorlinks=true,allcolors=blue!55!black]{hyperref}

\newcommand{\ntasks}{4}         %
\newcommand{\nscen}{2}          %
\newcommand{\nmat}{2}           %
\newcommand{\nbio}{2}           %
\newcommand{\nrunsfull}{288}    %
\newcommand{\nrunsmain}{144}
\newcommand{\nmodels}{11}        %
\newcommand{\nrunsall}{3168}      %
\newcommand{\nrunsallmain}{1584}  %
\newcommand{\benchname}{SkillBuffBench}
\newcommand{\cnone}{\ensuremath{\varnothing}}
\newcommand{\ccur}{\ensuremath{\mathrm{S}}}
\newcommand{\cauto}{\ensuremath{\mathrm{A}}}
\newcommand{\mzero}{\ensuremath{\mathrm{M0}}}
\newcommand{\mone}{\ensuremath{\mathrm{M1}}}

\title{\benchname: Measuring What Agent Skills Contribute\\
across Harnesses, Models, Skill Access Modes and Memory}

\author{%
  Gengyue Dong$^{1,*}$ \quad Ye Wei$^{2,*}$ \quad Zheng Zhu$^{3,*}$\\[4pt]
  Yidong Han$^{3,*}$ \quad Jiale Miao$^{4,*}$ \quad
  Guanghui Li$^{5,\dagger}$\\[8pt]
  {\small $^{1}$Department of Engineering Science, University of Oxford,
          Oxford, UK}\\
  {\small $^{2}$MRC Weatherall Institute of Molecular Medicine, University of
          Oxford, Oxford, UK}\\
  {\small $^{3}$Department of Materials, Imperial College London, London, UK}\\
  {\small $^{4}$Department of Mathematics, University College London,
          London, UK}\\
  {\small $^{5}$BraneMatrix AI, Shanghai, China}\\[4pt]
  {\small $^{*}$Equal contribution. \quad $^{\dagger}$Corresponding author:
          \texttt{raymondli@branematrix.com}}%
}
\date{}

\definecolor{sbgreen}{HTML}{D5E5DC}
\definecolor{sbgreenD}{HTML}{7FAF95}
\definecolor{sbpeach}{HTML}{F8E3C8}
\definecolor{sbpeachD}{HTML}{DDA85F}
\definecolor{sblav}{HTML}{DCD3EA}
\definecolor{sblavD}{HTML}{9F8DC6}
\definecolor{sbgray}{HTML}{8C8C8C}
\definecolor{sbaccent}{HTML}{C0603A}
\newcommand{\sbtask}[2]{\par\noindent\hangindent=9pt\hangafter=1\relax
  \makebox[9pt][l]{#1.}#2\par}
\tikzset{
  sbpanel/.style={rounded corners=5pt, fill=#1, draw=none},
  sbcard/.style={rounded corners=3pt, fill=white, draw=black!22,
                 line width=0.4pt, align=center, inner sep=3pt,
                 font=\scriptsize},
  sbpill/.style={rounded corners=6pt, fill=white, draw=#1, line width=0.8pt,
                 align=center, inner sep=2.5pt, font=\scriptsize},
  sbfat/.style={-{Triangle[length=6pt,width=10pt]}, line width=5pt,
                draw=sbgray, shorten >=1pt, shorten <=1pt},
  sbarrow/.style={-{Latex[length=4pt,width=4pt]}, line width=0.7pt,
                  draw=sbgray},
  sbtitle/.style={font=\small\bfseries, anchor=north},
  sblabel/.style={font=\scriptsize\bfseries, anchor=north},
}

\begin{document}
\maketitle

\begin{abstract}
Agent Skills---file-based packages of procedural knowledge that a coding agent
can discover and load on demand---are now shipped by every major agent harness,
yet the evidence for how much they help comes almost entirely from settings in
which the right Skill is hand-picked for the task and the agent starts from a
blank slate. Neither holds in deployment, where an agent chooses from whatever
library happens to be installed and frequently carries memory from earlier
work. We present \benchname, a benchmark and paired evaluation protocol that
treats Skill access as a three-level factor (no Skill, a curated Skill, and
automatic invocation from an installed library), crosses it with memory (off
and on), with four agent harnesses and with multiple models (the state-of-the-art proprietary models from Anthropic,
Google and OpenAI, and the state-of-the-art open-weight models from Chinese
providers), every model being run under
every harness, and evaluates the result on research tasks contributed by domain experts together
with the Skills they use, currently in materials science and biology, with the
suite extended on a rolling basis. Every run
is marked by a deterministic verifier; uncertainty is reported with a paired,
task-level bootstrap; and cost, agent rounds and wall-clock time are reported
alongside the score rather than folded into it. This version of the paper
describes the design of the benchmark, its evaluation protocol and its current
tasks and Skills. The evaluation is in progress: results will be added once
the runs are complete, and the task suite, results and analysis will continue
to be updated on a rolling basis.
\end{abstract}

\section{Introduction}
\label{sec:introduction}

\begin{figure}[t]
  \centering
  \resizebox{\linewidth}{!}{\begin{tikzpicture}[x=1.2cm,y=1.2cm,
    every node/.append style={execute at begin node={\hyphenpenalty=10000\relax\exhyphenpenalty=10000\relax}}]
  \path[sbpanel=sbgreen] (0,0)     rectangle (4.9,-7.0);
  \path[sbpanel=sbpeach] (5.45,0)  rectangle (10.35,-7.0);
  \path[sbpanel=sblav]   (10.9,0)  rectangle (15.8,-7.0);
  \draw[sbfat] (4.87,-3.5) -- (5.50,-3.5);
  \draw[sbfat] (10.32,-3.5) -- (10.95,-3.5);

  \node[sbtitle] at (2.45,-0.15) {Task suite};
  \node[font=\scriptsize, anchor=north] at (2.45,-0.65)
       {\ntasks{} expert-contributed tasks, \nscen{} domains};
  \node[sbcard, text width=4.93cm, inner xsep=4pt, align=left, anchor=north,
        draw=sbgreenD] (mat) at (2.45,-1.2)
       {{\centering\textbf{Materials science (\nmat)}\par}
        \sbtask{1}{oxygen tracer diffusion from SIMS}
        \sbtask{2}{XRD phase identification and refinement}};
  \node[sbcard, text width=4.93cm, inner xsep=4pt, align=left, anchor=north,
        draw=sbgreenD] (bio) at ([yshift=-0.2cm]mat.south)
       {{\centering\textbf{Biology (\nbio)}\par}
        \sbtask{3}{regulatory target prioritisation}
        \sbtask{4}{spheroid structure from 3D microscopy}};
  \node[font=\scriptsize\itshape, anchor=north] (tsnoteA)
       at ([yshift=-0.08cm]bio.south)
       {each task comes with its expert's Skill};
  \node[font=\scriptsize\itshape, anchor=north, inner ysep=1pt] (tsnoteB)
       at ([yshift=0.06cm]tsnoteA.south)
       {suite extended on a rolling basis};
  \node[sbcard, text width=5.0cm, align=left, anchor=north, inner sep=5pt]
       at ([yshift=-0.14cm]tsnoteB.south)
       {\textbf{Given to the agent}\\
        task prompt $\cdot$ input assets\\
        pinned toolset $\cdot$ fresh sandbox\\[4pt]
        \textbf{Hidden from the agent}\\
        reference solution $\cdot$ golden values\\
        deterministic verifier};

  \node[sbtitle] at (7.9,-0.15) {Paired factorial design};
  \node[sblabel] at (7.9,-0.68) {4 harnesses};
  \node[sbcard, minimum width=2.6cm] at (6.725,-1.3) {Ginkley};
  \node[sbcard, minimum width=2.6cm] at (9.075,-1.3) {Codex};
  \node[sbcard, minimum width=2.6cm] at (6.725,-1.8) {Claude Code};
  \node[sbcard, minimum width=2.6cm] at (9.075,-1.8) {OpenCode};
  \node[font=\scriptsize, anchor=north] at (7.9,-2.12)
       {$\times$ multiple models, each under every harness};

  \node[sblabel] at (7.9,-2.7) {$\times$ 3 Skill conditions};
  \node[sbpill=sbgreenD, minimum width=1.72cm, minimum height=1.0cm]
       at (6.36,-3.6) {$\varnothing$\\no Skill};
  \node[sbpill=sbpeachD, minimum width=1.72cm, minimum height=1.0cm]
       at (7.9,-3.6) {S\\curated};
  \node[sbpill=sblavD, minimum width=1.72cm, minimum height=1.0cm]
       at (9.44,-3.6) {A\\automatic};

  \node[sblabel] at (7.9,-4.4) {$\times$ 2 memory modes};
  \node[sbcard, minimum width=2.6cm, minimum height=0.95cm] at (6.725,-5.28)
       {M0\\memory off};
  \node[sbcard, minimum width=2.6cm, minimum height=0.95cm] at (9.075,-5.28)
       {M1\\memory on};
  \node[sblabel] at (7.9,-5.95) {$\times$ 3 repeats per cell};
  \node[font=\scriptsize\itshape, anchor=north] at (7.9,-6.35)
       {same task, toolset and budget};

  \node[sbtitle] at (13.35,-0.15) {Verification and reporting};
  \node[sbcard, text width=5.0cm, anchor=north] (ver) at (13.35,-0.8)
       {\textbf{Deterministic verifier}\\
        metrics specific to each task\\
        (residuals, $R_{\mathrm{wp}}$, average precision)\\
        normalised to a continuous\\
        score $y\in[0,1]$ $\cdot$ pass if $y\ge\tau_i$};
  \node[sbcard, text width=5.0cm, anchor=north] (eff)
       at ([yshift=-0.5cm]ver.south)
       {\textbf{Effect of Skills}\\[2pt]
        $\Delta_{\mathrm{S}} = P_{\mathrm{S}} - P_{\varnothing}$\qquad
        $\Delta_{\mathrm{A}} = P_{\mathrm{A}} - P_{\varnothing}$\\[2pt]
        normalised gain $g$\\
        paired bootstrap 95\% CI};
  \draw[sbarrow] (ver.south) -- (eff.north);
  \node[sbcard, text width=5.0cm, anchor=north] (mem)
       at ([yshift=-0.45cm]eff.south)
       {\textbf{Effect of memory}\\
        M1 against M0, Skill condition fixed};
  \node[sbcard, text width=5.0cm, anchor=north] (also)
       at ([yshift=-0.45cm]mem.south)
       {\textbf{Reported alongside}\\
        reliability: pass$^{k}$, pass@$k$\\
        cost, tokens, rounds, time\\
        failure types\\
        Skill invocation rate};
\end{tikzpicture}}
  \caption{\textbf{\benchname{} at a glance.} Expert-contributed research
  tasks, currently \ntasks{} in materials science and biology, are run under a paired factorial design that crosses
  four harnesses and multiple models, each model run under every harness,
  with three Skill conditions and two memory modes. A deterministic verifier marks every run; the effect of
  Skills and of memory is reported with paired bootstrap confidence
  intervals, next to reliability, efficiency and failure types.}
  \label{fig:overview}
\end{figure}
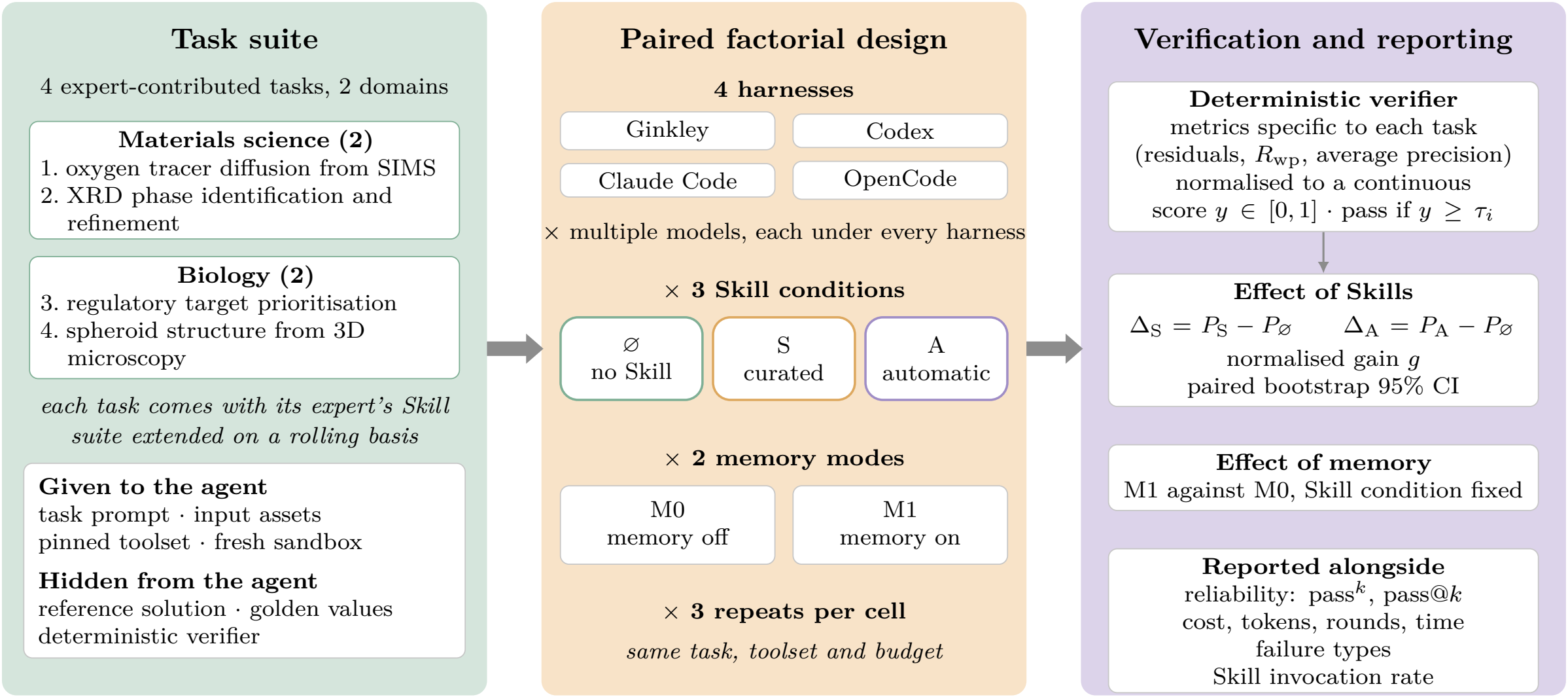

Language-model agents are increasingly judged not by what the underlying model
knows but by what the system around it lets the model do. A large part of that
system is now written in plain files. An \emph{Agent
Skill}~\citep{anthropic2025skills} is a directory containing a short
description, a set of instructions and, optionally, scripts, templates and
reference material; the agent \emph{harness}---the program that runs the tool
loop, manages the workspace and decides what enters the context
window---advertises the descriptions of the installed Skills to the model and
loads the body of one only when the model asks for it. The idea of giving an
agent a reusable library of procedures is older than the
format~\citep{wang2024voyager}, but the format has made such libraries cheap to
write and easy to share. An agent in deployment now commonly runs with a
library of Skills installed, most of them irrelevant to the task at hand.

Whether those Skills help is an empirical question, and it has only recently
been asked carefully. Benchmarks for agents now cover a wide range of work
\citep{yehudai2026survey}: resolving software issues
\citep{jimenez2024swebench,chowdhury2024sweverified}, operating a terminal
\citep{merrill2026terminalbench}, completing workplace and customer-service
tasks through tools \citep{styles2024workbench,yao2024taubench,barres2025tau2},
calling functions correctly
\citep{patil2025bfcl,chen2025acebench,zhong2025complexfuncbench}, browsing and
research \citep{mialon2024gaia,chen2026browsecompplus}, using a desktop
\citep{xie2024osworld}, planning \citep{dagan2025plancraft}, and domain work
in finance and civil engineering
\citep{bigeard2025financeagent,li2025drafterbench}. A parallel line measures
what agents can do in cybersecurity, from capture-the-flag challenges to
finding, exploiting and patching vulnerabilities in real codebases
\citep{zhang2025cybench,zhang2025bountybench,wang2026cybergym,shi2026cybergyme2e,zhang2026mobilecybench},
and another measures how long a task an agent can sustain
\citep{kwa2025horizon} and whether it holds to its deployment policy under
attack \citep{zou2025art,dziemian2026ipi}. These benchmarks share a protocol:
a sandboxed environment, a well-defined task, a deterministic check of the
final state, and a comparison across models and, sometimes, across harnesses,
tool sets or retrieval methods. What they hold constant is the procedural
knowledge available to the agent. They vary the agent and fix its
augmentation, and, to our knowledge, none treats persistent memory as an
experimental variable.
SkillsBench~\citep{li2026skillsbench} inverted this, holding the task, model
and harness fixed and varying only whether curated Skills are available. Across
87 tasks and 18 model--harness configurations it found that curated Skills
raise the mean score by 16.6 percentage points, that the gain varies widely by
domain and by harness, that 13 tasks get \emph{worse} with Skills, and that
Skills written by the model itself did not substitute for curated ones. That
study establishes that Skills can matter. It leaves open three questions that
decide whether they matter in use.

\paragraph{Curated access is an upper bound, not a deployment condition.}
In a curated-Skill condition a person who knows the task has already solved the
hardest part of the problem: deciding which knowledge is relevant. A deployed
agent receives no such help. It sees the descriptions of every installed
Skill, must recognise that one of them applies, must choose it over plausible
distractors, and must then follow it. Each step can fail, and loading the
wrong Skill can be worse than loading none. The gap between \emph{the right
Skill, supplied} and \emph{a library, searched} is the quantity a practitioner
needs, and it cannot be read off a curated-versus-none comparison.

\paragraph{Memory is an uncontrolled second source of procedural knowledge.}
Modern harnesses persist information between sessions: project instructions,
notes the agent wrote for itself, records of what worked last time. Memory and
Skills overlap in function---both carry procedure into a fresh context---so an
experiment that does not control memory cannot attribute an improvement to
either. Worse, repeated trials of one task are no longer independent if the
second trial can read what the first one learned. Whether a Skill still helps
an agent that already remembers a successful attempt, and whether memory alone
recovers the benefit of a Skill, are open questions.

\paragraph{Model and harness effects are confounded.}
Skill discovery and loading are implemented by the harness, not the model, and
harnesses implement them differently. Most existing evidence pairs each model
with its vendor's own harness, or varies the model inside a single harness, so
a difference between two rows of a results table may be due to the model, the
harness, or their interaction. Separating them requires running the
\emph{same} models under \emph{different} harnesses.

\medskip
This paper introduces \benchname, a benchmark and evaluation protocol built to
answer these three questions. Its design is a full factorial: three Skill
conditions (none, curated, and automatic invocation from an installed
library), two memory modes (off and on), four harnesses, and multiple models
from different providers, each run under every harness, with every cell evaluated on the same tasks and repeated three times. The
tasks are research problems contributed by domain experts, each with the
Skill that expert uses for it. They currently come from materials science and
biology: extracting oxygen transport coefficients from isotope-exchange
depth profiles, identifying a crystalline phase and refining its unit cell
from a powder diffraction pattern, prioritising candidate regulatory
targets of a gene for experimental follow-up, and measuring the structure of
a cell spheroid and the distribution of its cells from three-dimensional
microscopy images. Such workflows depend on
procedural knowledge that is specialised, written down somewhere, and
unevenly represented in what models already know, and they require
professional software and data rather than a general-purpose programming
environment. The suite is extended on a rolling basis.

Every run is marked by a deterministic verifier against versioned inputs, so
that scores are reproducible and comparable across harnesses.

\paragraph{Contributions.}
\begin{enumerate}[leftmargin=*,itemsep=2pt,topsep=4pt]
  \item \textbf{A deployment-facing Skill condition.} Alongside the no-Skill
        baseline and the curated-Skill condition, we evaluate automatic
        invocation from an installed Skill library, which separates the value
        of a Skill's \emph{content} from the cost of \emph{finding} it.
  \item \textbf{Memory as a controlled factor.} Every Skill condition is run
        with memory disabled and strictly isolated, and again with memory
        enabled, so that the contribution of Skills is not confounded with
        that of memory and the interaction between the two can be measured.
  \item \textbf{A crossed harness-by-model design.} Four harnesses are driven
        by a common pool of models with a task-defined toolset, so harness
        effects are compared only at a fixed model.
  \item \textbf{Expert-contributed tasks with deterministic verifiers.} The
        tasks and their Skills come from practising researchers, currently in
        materials science and biology, each with pinned tools, versioned
        inputs and a deterministic verifier.
  \item \textbf{Statistical and efficiency reporting.} We report absolute and
        normalised gains with paired task-level bootstrap confidence
        intervals, and cost, agent rounds and wall-clock time next to every
        score.
\end{enumerate}

\paragraph{Status of this version.}
The evaluation is in progress at the time of writing. This version specifies
the design (Section~\ref{sec:methodology}) and introduces the current tasks
and Skills (Section~\ref{sec:suite}). Results will be added once the runs are
complete. The paper is a living document: further tasks, results, analysis
and a full discussion of related work will be added on a rolling basis, and
the counts reported here describe the suite as it stands.

\section{Methodology}
\label{sec:methodology}

\subsection{Research questions and design overview}
\label{sec:overview}

\benchname{} measures whether Skills raise task success for agents working in
realistic conditions, and whether that effect survives changes of harness,
task domain and memory setting. Three questions organise the design:

\begin{description}[leftmargin=2.6em,labelwidth=2.1em,itemsep=1pt,topsep=4pt]
  \item[RQ1] Does a \emph{curated} Skill improve task success over no Skill?
  \item[RQ2] Does \emph{automatic invocation} from an installed Skill library
             improve task success over no Skill, and how much of the curated
             gain does it retain?
  \item[RQ3] Does memory change task success, cost, number of agent rounds or
             wall-clock time, and does it change the effect of Skills?
\end{description}

\begin{figure}[t]
  \centering
  \resizebox{\linewidth}{!}{\begin{tikzpicture}[x=1cm,y=1cm,
    every node/.append style={execute at begin node={\hyphenpenalty=10000\relax\exhyphenpenalty=10000\relax}}]
  \node[font=\scriptsize\bfseries, anchor=north west] at (0,0.45) {(a) One task, three Skill conditions};

  \node[rounded corners=4pt, draw=black!30, line width=0.5pt, fill=white,
        text width=3.45cm, align=left, anchor=north west, inner sep=5pt,
        font=\scriptsize] (task) at (0,0)
       {{\footnotesize\textbf{\benchname{} task}}\\[3pt]
        \textbf{Given to the agent}\\
        \textcolor{sbgray}{\tiny$\blacksquare$} task prompt\\
        \textcolor{sbgray}{\tiny$\blacksquare$} input assets\\
        \textcolor{sbgray}{\tiny$\blacksquare$} pinned toolset\\
        \textcolor{sbgray}{\tiny$\blacksquare$} fresh sandbox\\[3pt]
        \textbf{Varied}\\
        \textcolor{sbgray}{\tiny$\blacksquare$} Skill access\\
        \textcolor{sbgray}{\tiny$\blacksquare$} memory (M0 / M1)\\[3pt]
        \textbf{Withheld}\\
        \textcolor{sbgray}{\tiny$\blacksquare$} reference solution\\
        \textcolor{sbgray}{\tiny$\blacksquare$} verifier, golden values};

  \node[rounded corners=4pt, fill=sbgreen, text width=4.7cm, align=left,
        anchor=west, inner sep=5pt, minimum height=1.15cm, font=\scriptsize]
        (c0) at (4.75,-0.75)
       {\textbf{Condition $\varnothing$ $\cdot$ no Skill}\\
        \textit{prompt and toolset only}};
  \node[rounded corners=4pt, fill=sbpeach, text width=4.7cm, align=left,
        anchor=west, inner sep=5pt, minimum height=1.15cm, font=\scriptsize]
        (c1) at (4.75,-2.25)
       {\textbf{Condition S $\cdot$ curated Skill}\\
        \textit{relevant Skills chosen by a person}};
  \node[rounded corners=4pt, fill=sblav, text width=4.7cm, align=left,
        anchor=west, inner sep=5pt, minimum height=1.15cm, font=\scriptsize]
        (c2) at (4.75,-3.75)
       {\textbf{Condition A $\cdot$ auto-invocation}\\
        \textit{installed library; the harness selects}};

  \coordinate (hub) at (4.15,-2.25);
  \draw[line width=1.3pt, sbgreenD] (task.east |- hub) -- (hub)
        to[out=0,in=180] (c0.west);
  \draw[line width=1.3pt, sbpeachD] (task.east |- hub) -- (hub) -- (c1.west);
  \draw[line width=1.3pt, sblavD]   (task.east |- hub) -- (hub)
        to[out=0,in=180] (c2.west);

  \node[sbpill=sbgreenD, minimum width=1.05cm, minimum height=0.55cm]
        (p0) at (11.15,-0.75) {$P_{\varnothing}$};
  \node[sbpill=sbpeachD, minimum width=1.05cm, minimum height=0.55cm]
        (p1) at (11.15,-2.25) {$P_{\mathrm{S}}$};
  \node[sbpill=sblavD, minimum width=1.05cm, minimum height=0.55cm]
        (p2) at (11.15,-3.75) {$P_{\mathrm{A}}$};
  \draw[line width=1.3pt, sbgreenD] (c0.east) -- (p0.west);
  \draw[line width=1.3pt, sbpeachD] (c1.east) -- (p1.west);
  \draw[line width=1.3pt, sblavD]   (c2.east) -- (p2.west);

  \node[anchor=west, align=left, font=\scriptsize] (g1) at (12.55,-1.5)
       {{\tiny\sffamily\bfseries\textcolor{sbaccent}{CURATED GAIN}}\\[1pt]
        $\Delta_{\mathrm{S}} = P_{\mathrm{S}} - P_{\varnothing}$};
  \node[anchor=west, align=left, font=\scriptsize] (g2) at (12.55,-3.0)
       {{\tiny\sffamily\bfseries\textcolor{sbaccent}{AUTOMATIC GAIN}}\\[1pt]
        $\Delta_{\mathrm{A}} = P_{\mathrm{A}} - P_{\varnothing}$};
  \draw[line width=0.7pt, sbgray] (p0.east) to[out=0,in=180] (g1.west);
  \draw[line width=0.7pt, sbgray] (p1.east) to[out=0,in=180] (g1.west);
  \draw[line width=0.7pt, sbgray, dashed] (p0.east) to[out=0,in=180] (g2.west);
  \draw[line width=0.7pt, sbgray, dashed] (p2.east) to[out=0,in=180] (g2.west);

  \node[font=\scriptsize\itshape, anchor=north west, text width=10.9cm,
        align=left] at (4.75,-4.5)
       {Every condition is run with memory off (M0, headline) and on (M1),
        under each of 4 harnesses and each model of the shared pool, 3 times.};

  \node[font=\scriptsize\bfseries, anchor=north west] at (0,-5.55) {(b) One run};

  \node[rounded corners=4pt, fill=sbgreen, text width=2.45cm, align=center,
        anchor=north west, inner sep=4pt, minimum height=1.55cm,
        font=\scriptsize] (s1) at (0,-6.0)
       {\textbf{1\; Reset}\\new workspace, session, memory store};
  \node[rounded corners=4pt, fill=sbgreen, text width=2.45cm, align=center,
        anchor=north west, inner sep=4pt, minimum height=1.55cm,
        font=\scriptsize] (s2) at (3.2625,-6.0)
       {\textbf{2\; Agent run}\\harness and model under fixed budgets};
  \node[rounded corners=4pt, fill=sbpeach, text width=2.45cm, align=center,
        anchor=north west, inner sep=4pt, minimum height=1.55cm,
        font=\scriptsize] (s3) at (6.525,-6.0)
       {\textbf{3\; Collect}\\artifacts, logs, tokens, cost, time};
  \node[rounded corners=4pt, fill=sblav, text width=2.45cm, align=center,
        anchor=north west, inner sep=4pt, minimum height=1.55cm,
        font=\scriptsize] (s4) at (9.7875,-6.0)
       {\textbf{4\; Verify}\\reference values, withheld labels, artifacts};
  \node[rounded corners=4pt, fill=sblav, text width=2.45cm, align=center,
        anchor=north west, inner sep=4pt, minimum height=1.55cm,
        font=\scriptsize] (s5) at (13.05,-6.0)
       {\textbf{5\; Aggregate}\\repeats $\rightarrow$ tasks $\rightarrow$ $P$, with bootstrap CI};
  \draw[sbarrow, line width=1pt] (s1.east) -- (s2.west);
  \draw[sbarrow, line width=1pt] (s2.east) -- (s3.west);
  \draw[sbarrow, line width=1pt] (s3.east) -- (s4.west);
  \draw[sbarrow, line width=1pt] (s4.east) -- (s5.west);
\end{tikzpicture}}
  \caption{\textbf{Evaluation protocol.} (a)~Each task is run under three
  Skill conditions that differ only in Skill access. The macro-average scores
  $P_{\varnothing}$, $P_{\mathrm{S}}$ and $P_{\mathrm{A}}$ give the curated
  gain $\Delta_{\mathrm{S}}$ and the automatic gain $\Delta_{\mathrm{A}}$
  against the same baseline. (b)~Every run starts from a reset environment,
  and the verifier sees only what the run leaves behind.}
  \label{fig:protocol}
\end{figure}
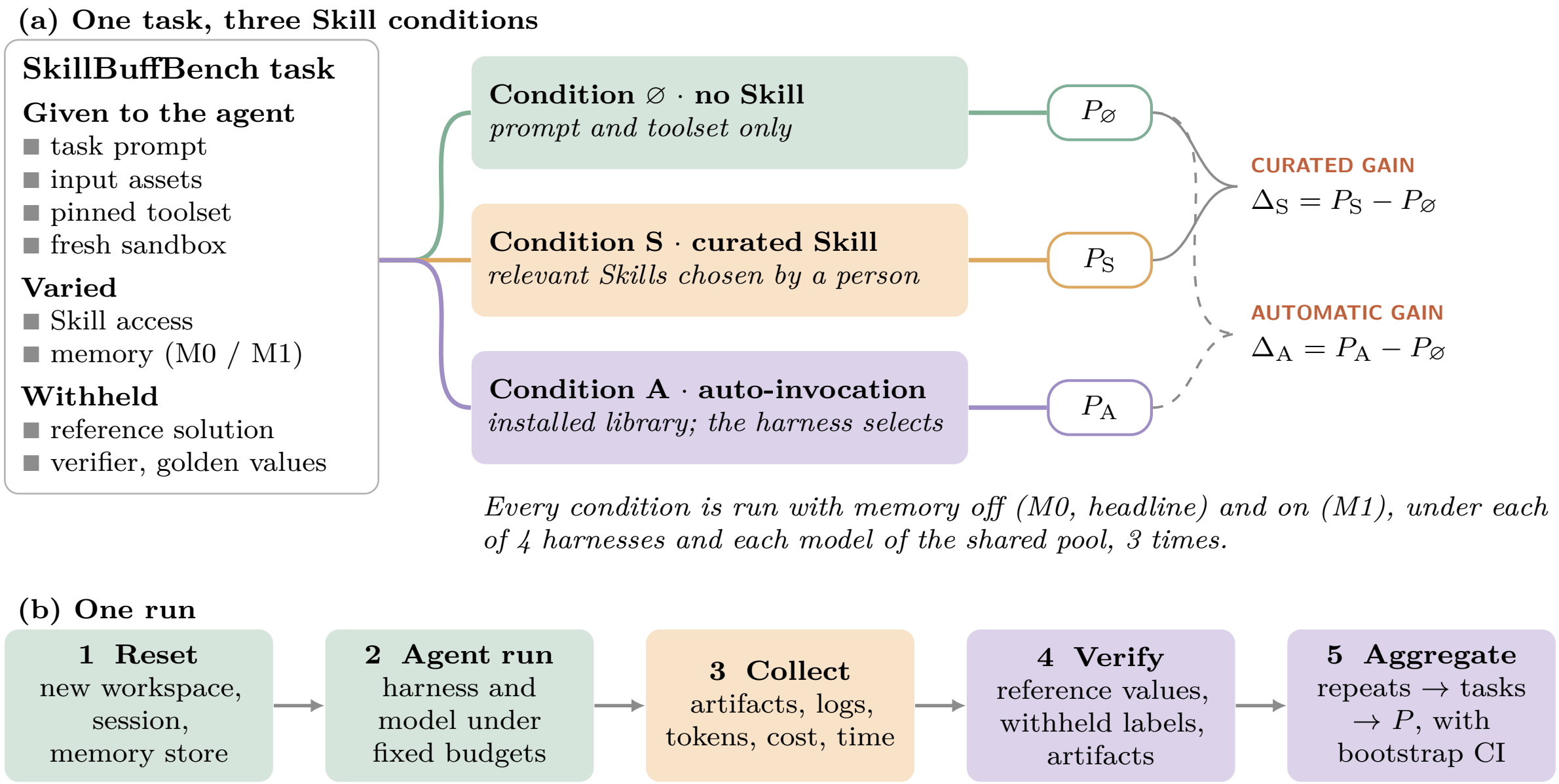

The experiment is a full factorial over four factors, summarised in
Table~\ref{tab:factors} and Figure~\ref{fig:protocol}, evaluated on a fixed suite of \ntasks{} tasks with three
repeats per cell. A \emph{configuration} is a (harness, model) pair; within a
configuration, the only things that vary between the runs of a task are the
Skill condition and the memory mode. All comparisons that support RQ1--RQ3
are therefore paired on task, harness and model.

\begin{table}[t]
  \centering
  \small
  \caption{Experimental factors. ``Levels'' is the number of values a factor
  takes; $n$ is the size of the model pool. Each cell is evaluated on all
  \ntasks{} tasks with three repeats.}
  \label{tab:factors}
  \begin{tabularx}{\linewidth}{@{}llX@{}}
    \toprule
    Factor & Levels & Values \\
    \midrule
    Harness         & 4 & Ginkley, Codex, Claude Code, OpenCode \\
    Model           & $n$ (now \nmodels) & one pool shared by all harnesses, updated as
                      new state-of-the-art models are released
                      (Section~\ref{sec:harnesses}) \\
    Skill condition & 3 & \cnone{} no Skill;\; \ccur{} curated Skill;\;
                      \cauto{} automatic invocation from an installed library \\
    Memory mode     & 2 & \mzero{} memory off, isolated;\; \mone{} memory on \\
    \bottomrule
  \end{tabularx}
\end{table}

\subsection{Harnesses and models}
\label{sec:harnesses}

We evaluate agents as they are used: a model inside a harness. The four
harnesses are Ginkley, developed by BraneMatrix; Codex, by OpenAI; Claude
Code, by Anthropic; and OpenCode, an open-source project. Each is run at its
latest release at the time of the runs, and the exact version is recorded
with every result.

The harnesses handle Skills in the same broad way, which follows the open
Agent Skills format~\citep{agentskills2025spec,anthropic2025skills}. A Skill
is a directory containing a \texttt{SKILL.md} file whose header gives the
Skill's name and a short description. When a session starts, the harness
scans its Skill directories and places only these names and descriptions in
the model's context. The model itself decides, from the descriptions, whether
a Skill applies to the task at hand; only then is the body of that Skill
loaded, and any scripts or reference files it bundles are read as they are
needed~\citep{claudecode2026skills,codex2026skills,opencode2026skills}.
Selection is thus made by the model, but on terms set by the harness: which
directories are scanned, how the descriptions are presented, and how many fit
in the context. Memory follows a common pattern as well. Each harness reads
persistent instruction files from the project and from the user's
configuration (\texttt{CLAUDE.md} in Claude Code, \texttt{AGENTS.md} in Codex
and OpenCode) into the context at the start of every session, and Claude
Code in addition keeps notes that the agent writes for itself and reloads in
later sessions~\citep{claudecode2026memory,codex2026agentsmd,opencode2026rules}.

Rather than pairing each harness with its vendor's
model, every harness is driven by the same pool of models. The pool has two
groups: the state-of-the-art proprietary models of the three major
international providers (Anthropic, Google with its Gemini family, and
OpenAI), and the state-of-the-art open-weight models released by Chinese
providers. From each international provider we take its two most capable
generally available models, and from each Chinese provider its single most
capable open-weight model. We write $n$ for the size of the pool; it
currently holds $n = \nmodels$ models (Table~\ref{tab:models}). The pool is
not fixed: as new state-of-the-art models are released they are added to it
and run under the same protocol, so $n$ changes with successive versions of
the benchmark, and each version of the results states the pool it was run on.

\begin{table}[t]
  \centering
  \small
  \caption{The model pool at the time of writing ($n = \nmodels$). Two models
  from each international provider and one open-weight model from each
  Chinese provider.}
  \label{tab:models}
  \begin{tabular}{@{}lll@{}}
    \toprule
    Group & Provider & Models \\
    \midrule
    Proprietary & Anthropic & Claude Fable 5.2; Claude Opus 5.5 \\
                & Google    & Gemini 4 Argon; Gemini 3.8 Flash \\
                & OpenAI    & GPT-6 Astra; GPT-6.1 Sol \\
    \addlinespace
    Open-weight & DeepSeek    & DeepSeek-V4-Pro \\
                & Alibaba     & Qwen3.8 \\
                & Moonshot AI & Kimi K3.1 Pro \\
                & Z.ai        & GLM-5.3 \\
                & MiniMax     & MiniMax-M3 \\
    \bottomrule
  \end{tabular}
\end{table}

Results are reported per harness--model pair. Comparisons \emph{between}
harnesses are made only at a fixed model, and comparisons between models only
at a fixed harness, so that a model difference is never read as a harness
difference or the reverse.

\subsection{Skill conditions}
\label{sec:skill-conditions}

Each task is run under three Skill conditions inside each configuration.

\begin{description}[leftmargin=1.6em,itemsep=2pt,topsep=4pt]
  \item[\cnone{} --- No Skill.] No Skill is installed. The agent completes
        the task with the harness's default capabilities and the task's
        toolset. This is the baseline.
  \item[\ccur{} --- Curated Skill.] The Skills most relevant to the task,
        selected by a person, are made available. This isolates the value of
        Skill \emph{content} when the selection problem has been solved for
        the agent.
  \item[\cauto{} --- Automatic invocation.] The agent runs with a Skill
        library installed, as it would in production, and the harness's own
        mechanism decides which Skills, if any, are retrieved and invoked.
        This adds the cost of discovery and the risk of selecting a wrong or
        irrelevant Skill.
\end{description}

The difference between \ccur{} and \cnone{} bounds what Skills can
contribute; the difference between \cauto{} and \ccur{} is what is lost to
discovery. In all three conditions the task
prompt is identical and does not name any Skill.

Skills are held to the same leakage rule as in prior
work~\citep{li2026skillsbench}: a Skill must serve a class of problems. It may
not contain the output of a particular task, task-specific paths or constants,
or commands that solve one instance directly. Reference solutions, golden
values and verifier code are never visible to the agent.

\paragraph{What is disclosed about Skills.}
Several of the Skills evaluated here were contributed by collaborators and
remain their property. This paper therefore describes the benchmark tasks in
full and each Skill only by its \emph{purpose}: the class of problems it
serves. The text of a Skill, its instructions, scripts, templates and
reference material are not reproduced.
Section~\ref{sec:suite} gives that purpose for each contributed Skill.

\subsection{Memory modes}
\label{sec:memory}

Every Skill condition is run under two memory modes.

\begin{description}[leftmargin=1.6em,itemsep=2pt,topsep=4pt]
  \item[\mzero{} --- Memory off.] Each run starts as a new agent: a new
        session, a new working directory and a new memory namespace.
  \item[\mone{} --- Memory on.] The agent may use memory of successful task
        completions accumulated under the same experimental definition.

\end{description}

Headline results use \mzero{}. \mone{} is reported separately, as a contrast
that quantifies what memory itself contributes and how it interacts with
Skills (RQ3). We make \mzero{} the default because it is the only mode in
which the three repeats of a cell are exchangeable and in which an improvement
can be attributed to the Skill condition alone.

\paragraph{Isolation procedure.}
To make ``memory off'' true in practice, each \mzero{} run is launched under
the following rules.
\begin{enumerate}[leftmargin=*,itemsep=1pt,topsep=3pt]
  \item A new temporary workspace is created for the run.
  \item The session is never resumed from, or continued into, another session.
  \item History directories, cache directories and memory namespaces are not
        shared between runs.
  \item General memory is disabled or, where a harness cannot disable it, the
        run is given a fresh namespace.
  \item After the run, only artifacts, logs and metrics are collected.
        Nothing a run produced is written back into the context of a later run.
\end{enumerate}
API credentials may be reused. What may not be reused is session history,
working-directory state, the memory namespace, and any project memory that the
harness loads automatically.

\subsection{Task suite}
\label{sec:tasks}

The tasks are contributed by domain experts among the authors, each together
with the Skill that expert uses for it. Section~\ref{sec:suite} introduces the
tasks and Skills; this subsection states the rules that every task follows,
whatever its domain.

\paragraph{Verifier families.}
Across the suite, verifiers take one of three forms, and a task may combine
them. \emph{Reference-value} verifiers compare quantities extracted from the
agent's output with values from a reference solution, within a tolerance
stated in the task. \emph{Withheld-label} verifiers score a frozen prediction
against outcomes the agent never saw, using a metric fixed before the run.
\emph{Artifact and behaviour} verifiers check that the required files exist
and are well formed, and that the agent's report makes the claims the
evidence supports and no others, including declining to conclude when the
evidence is insufficient. All three are deterministic: no verifier relies on
a language model's judgement. A verifier need not return a bare pass or
fail. Where the result of a task is a continuous quantity, its verifier
returns a graded score (Section~\ref{sec:metrics}).

\paragraph{Leakage.}
Tasks built on published data carry the risk that an agent with network
access retrieves the answer instead of producing it. For such tasks the
outcomes are separated from the inputs before the run and the agent's runtime
excludes the outcome files. Network access is restricted as well. By default
a run has no access to the open internet: the sandbox allows outbound
connections only to the model's API. A task may declare the external
services its work requires, such as a crystal-structure database, and only
those are opened, for every harness and condition alike. The publication a
task is drawn from, and the repositories that hold its data, are never among
them. Every network request a run makes is recorded in its trajectory; a run
that attempts to reach a blocked source is flagged and reported, and its
result is not counted.

\paragraph{Task package.}
A task consists of (i) a prompt, written by a person, that states the goal and
the artifacts to leave behind; (ii) the input assets the agent receives;
(iii) a declared toolset with pinned versions; (iv) a reference solution that
the verifier must accept; (v) golden values derived from the reference
solution; and (vi) a deterministic verifier. Items (iv)--(vi) are never shown
to the agent. Task inputs, tool versions, data sources and verifier rules are
versioned, and their cryptographic hashes are recorded with every result so
that a score can be traced to the exact inputs that produced it and so that
inputs cannot drift between runs.

\paragraph{Toolsets belong to tasks, not harnesses.}
Each task declares the tools it permits and their versions. Every harness is
mapped onto that same set of equivalent tool entry points, and a tool that
only one harness offers is not enabled. Without this rule a harness with a
richer built-in toolset would appear to benefit less from Skills simply
because it needed them less.

\subsection{Inputs and recorded outputs}
\label{sec:io}

Agent benchmarks are most easily compared by what they vary and what they
measure. Table~\ref{tab:io} states both for \benchname. On the input side we
separate the variables we manipulate from what the agent is given and from
what is held fixed; on the output side we separate what is recorded for every
run from what is reported in aggregate.

\begin{table}[t]
  \centering
  \small
  \caption{Inputs and outputs of \benchname.}
  \label{tab:io}
  \begin{tabularx}{\linewidth}{@{}lX@{}}
    \toprule
    \multicolumn{2}{@{}l}{\textbf{Inputs}} \\
    \midrule
    Independent variables & harness; model; Skill condition
        (\cnone, \ccur, \cauto); memory mode (\mzero, \mone) \\
    Given to the agent & task prompt; input assets (documents, datasets,
        starter files); the task's declared toolset; a fresh, resettable
        sandbox workspace; under \ccur{} or \cauto, the Skills or Skill
        library; under \mone, the memory store \\
    Held fixed & model version; sampling parameters; output-token limit;
        budgets on agent rounds, wall-clock time and tokens; tool versions;
        task order; verifier \\
    Hidden from the agent & reference solution; golden values; verifier code \\
    \midrule
    \multicolumn{2}{@{}l}{\textbf{Outputs recorded for every run}} \\
    \midrule
    Outcome & verifier score in $[0,1]$; pass or fail against the task's threshold; result of each
        individual check \\
    Artifacts & the files the task requires, taken from the final state of
        the workspace \\
    Trajectory & the complete action log: messages, tool calls and their
        results, and Skill loads \\
    Skill usage & which Skills were offered, which were loaded, and which
        were invoked \\
    Resource use & input and output tokens; monetary cost; wall-clock time;
        agent rounds; tool calls; retries \\
    Diagnosis & primary failure type; boundary violations; infrastructure
        flag \\
    Provenance & model identifier and seed; hashes of the harness, toolset,
        task inputs and verifier \\
    \midrule
    \multicolumn{2}{@{}l}{\textbf{Outputs reported in aggregate}} \\
    \midrule
    Effectiveness & macro-average score $P$; absolute gain $\Delta$ and
        normalised gain $g$ for \ccur{} and \cauto; the memory contrast; all
        with bootstrap confidence intervals \\
    Reliability & pass$^k$ and pass@$k$ over the repeats of each task \\
    Efficiency & tokens, cost, time and rounds per run; cost per successful
        run \\
    Breakdown & per-domain and per-task scores; Skill invocation rate;
        distribution of failure types \\
    \bottomrule
  \end{tabularx}
\end{table}

Two choices in the table deserve comment. First, the trajectory is retained
and not only the final artifact. A verifier judges the end state, but whether
a Skill was found, read and followed, and whether the agent stayed inside the
task's boundaries, can only be established from what the agent did on the way
there. Second, Skill usage is recorded separately from the outcome. Under
\cauto{} a run can fail because the relevant Skill was never selected or
because it was selected and did not help, and these call for different
remedies; the score alone cannot tell them apart.

\subsection{Controlled variables}
\label{sec:controls}

Within a configuration, the following are held fixed across Skill conditions,
memory modes and repeats: the model name and version; the sampling parameters
(temperature, top-$p$ and seed, set to $0$, $1$ and a fixed seed wherever the
provider exposes them); the limit on output tokens; the toolset and tool
versions; the input assets, datasets, task prompt and verifier; the layout of
working, cache and memory directories; and the order in which tasks are run.
Every run starts from a freshly reset sandbox. Budgets on agent rounds,
wall-clock time and tokens are identical across conditions.
 The harness
version, toolset version and task-input version are recorded as hashes in each
run's configuration record.

\subsection{Run protocol}
\label{sec:protocol}

Every (task, harness, model, Skill condition, memory mode) cell is run three
times. For each model the full matrix is
\begin{equation*}
  4 \text{ harnesses} \times 3 \text{ Skill conditions} \times 2 \text{ memory
  modes} \times \ntasks \text{ tasks} \times 3 \text{ repeats} \;=\; \nrunsfull
  \text{ runs},
\end{equation*}
of which the $\nrunsmain$ runs under \mzero{} form the headline results. The
whole experiment is $n$ times this. With the current pool of $n = \nmodels$
models that is $\nmodels \times \nrunsfull = \nrunsall$ runs, of which
$\nrunsallmain$ are under \mzero.

\paragraph{Infrastructure failures.}
A run ends in one of three ways. If the verifier accepts or rejects the
result, the run is scored. If the run is interrupted by something outside the
agent's control---an API timeout, a network fault, a refused software licence
or a crashed runner---it is an infrastructure failure. Such runs are re-run or
flagged; they are never charged to the model. If infrastructure failures
recur for a cell, the cell is reported as not comparable.

\subsection{Metrics}
\label{sec:metrics}

\paragraph{Per-run outcome.}
Let $y_{ir}$ denote the verifier's outcome for task $i \in \{1,\dots,N\}$,
$N = \ntasks$, on repeat $r \in \{1,\dots,R\}$, $R = 3$, within one cell of the
design. We take $y_{ir} \in [0,1]$ to be the score assigned by the task's
deterministic verifier, and report alongside it the binary outcome
$b_{ir} = \mathbb{1}[y_{ir} \ge \tau_i]$, where $\tau_i$ is the pass threshold
of task $i$: a run passes if its score reaches the threshold and fails
otherwise. A graded outcome is compatible with
everything that follows, none of which requires $y_{ir}$ to be integral, and
it reduces the variance that makes three repeats uncomfortable: a
nearly-correct run contributes $0.7$ rather than swinging between $0$ and $1$.

\paragraph{Task-specific criteria, one common scale.}
The tasks do not share a notion of quality, and we do not force one on them.
What makes a diffusion fit good (small residuals between the fitted and the
measured profile, an interval that covers the true value) is not what makes a
phase identification or a ranking good. Each task therefore brings its own
criteria, chosen by the expert who contributed it, and the outcome of a run
is in general continuous and not a bare pass or fail. The only requirement
the benchmark imposes is that a task's verifier maps its criteria onto a
score in $[0,1]$ by a rule fixed before any run, where $1$ means that every
criterion is met in full. All scores then live on one scale, and the
aggregation, gains and intervals below apply to every task unchanged.
Section~\ref{sec:suite} states the criteria of each task.

Each task also has a pass threshold $\tau_i \in (0,1]$ on that score, set by
its contributor at the level below which a practitioner would not accept the
result, and fixed, like the mapping, before any run. The continuous score is
the primary outcome and is what the gains below are computed from; the
threshold turns it into the pass or fail that the reliability measures need.

\paragraph{Task score and macro average.}
The score of a task is the mean over its repeats, and the score of a cell is
the unweighted mean over tasks:
\begin{equation}
  p_i \;=\; \frac{1}{R}\sum_{r=1}^{R} y_{ir},
  \qquad
  P \;=\; \frac{1}{N}\sum_{i=1}^{N} p_i .
\end{equation}
The macro average gives every task equal weight, so a task does not count for
more because it is easier, cheaper or was run more often.

\paragraph{Gains.}
Write $P_{\cnone}$, $P_{\ccur}$ and $P_{\cauto}$ for the macro average under
the three Skill conditions of one configuration and memory mode. For
$c \in \{\ccur, \cauto\}$ we report the absolute gain and the normalised
gain~\citep{hake1998interactive},
\begin{equation}
  \Delta_c \;=\; P_c - P_{\cnone},
  \qquad
  g_c \;=\; \frac{P_c - P_{\cnone}}{1 - P_{\cnone}} .
\end{equation}
$\Delta_c$ is the improvement in percentage points; $g_c$ is the fraction of
the baseline's remaining headroom that the Skill condition recovers, which
makes configurations with different baselines comparable. The effect of
memory is reported analogously as the difference between \mone{} and \mzero{}
at a fixed Skill condition.

\paragraph{Uncertainty.}
Three repeats are too few for a standard error over repeats to be a reliable
primary estimate, and the tasks, not the repeats, are the main source of
variation. We therefore use a nonparametric bootstrap over
tasks~\citep{efron1979bootstrap}. For $b = 1,\dots,B$ with $B = 10{,}000$, we
draw $N$ task indices with replacement,
$I^{(b)} = (i_1,\dots,i_N)$, $i_j \sim \mathrm{Uniform}\{1,\dots,N\}$, and
recompute
\begin{equation}
  P^{(b)}_c \;=\; \frac{1}{N}\sum_{j=1}^{N} p_{c,\,i_j},
  \qquad
  \Delta^{(b)}_c \;=\; P^{(b)}_c - P^{(b)}_{\cnone},
\end{equation}
using the \emph{same} resampled index set for every condition, so that the
interval for a gain respects the pairing of conditions on tasks. The 95\%
confidence interval is the percentile interval
$[\,q_{2.5},\, q_{97.5}\,]$ of the bootstrap distribution; intervals for $g_c$
are obtained in the same way. The procedure resamples task scores that have
already been measured and involves no further agent runs.

\paragraph{Reliability.}
A mean over repeats does not distinguish an agent that succeeds on a task
every time from one that succeeds on a third of its attempts. Let
$b_{ir} = \mathbb{1}[y_{ir} \ge \tau_i]$ indicate that run $r$ of task $i$
passes its task's threshold. With
$k = R = 3$ we report
\begin{equation}
  \text{pass}^{k} \;=\; \frac{1}{N}\sum_{i=1}^{N}\,\prod_{r=1}^{k} b_{ir},
  \qquad
  \text{pass@}k \;=\; \frac{1}{N}\sum_{i=1}^{N}\,\max_{1\le r\le k} b_{ir},
\end{equation}
the fraction of tasks passed on \emph{all} $k$ attempts and on \emph{at
least one}~\citep{yao2024taubench}. The first is the relevant figure when a
failure is costly; the second bounds what retries could achieve; and the gap
between them measures how inconsistent the agent
is~\citep{rabanser2026reliability}. Comparing the gap across Skill conditions
shows whether Skills make success more repeatable as well as more frequent.

\paragraph{Efficiency.}
Resource use is not folded into the score. For each cell we report the mean
monetary cost per run, the mean number of tokens, the mean number of agent
decision rounds and the mean wall-clock time per run, and plot cost, rounds
and time against $P$. In all three plots the upper-left region is better: a
higher score at lower cost, in fewer rounds or in less time. We also report
the cost per successful run, the total cost of a cell divided by its number of
passing runs, which charges failed attempts to the successes they accompany.

\subsection{Failure taxonomy}
\label{sec:failures}

The score uses only the verifier outcome. For diagnosis, every failed run is
additionally given one primary failure type from the taxonomy in
Table~\ref{tab:failures}; secondary types may be recorded but do not enter any
statistic.

\begin{table}[t]
  \centering
  \small
  \caption{Failure taxonomy. Diagnostic only; it does not enter the score.}
  \label{tab:failures}
  \begin{tabularx}{\linewidth}{@{}lX@{}}
    \toprule
    Failure type & Meaning \\
    \midrule
    Task comprehension & misread the goal, missed a key requirement, or
        solved a different problem \\
    Tool use           & approach roughly right, but a tool call, command
        argument, script or software operation failed \\
    Data processing    & error in reading input, unit conversion, field
        matching, or handling of missing or anomalous values \\
    Reasoning / planning & task and tools understood, but wrong ordering of
        steps, experimental design or intermediate decision \\
    Output format      & result close to correct, but a required file, column
        name, schema or document structure is missing \\
    Verification       & no self-check, or too few checks run to confirm the
        artifact is correct \\
    Safety / boundary  & the agent acts outside the scope the task
        permits, or claims more than its evidence supports \\
    Infrastructure     & API, network, licence, dependency or runner failure
        unrelated to agent ability; excluded from scoring \\
    \bottomrule
  \end{tabularx}
\end{table}

\subsection{Reporting}
\label{sec:reporting}

For every harness--model pair we report: the macro average $P$ under each
Skill condition and memory mode; $\Delta_{\ccur}$, $\Delta_{\cauto}$,
$g_{\ccur}$ and $g_{\cauto}$ with bootstrap 95\% confidence intervals;
pass$^k$ and pass@$k$; the three efficiency plots and the cost per successful
run; per-domain and per-task scores; the Skill invocation rate under \cauto; the
distribution of failure types; and the versions of the model, tools, tasks, data and verifier that
produced the numbers. Per-task results are released in full, so that tasks on
which Skills hurt are visible and not averaged away.

\section{The \benchname{} Suite: Tasks and Skills}
\label{sec:suite}

This section introduces what the benchmark contains. Every task comes from
the research practice of a domain expert among the authors, and every task is
paired with the Skill that expert uses for it. The curated condition of
Section~\ref{sec:skill-conditions} therefore tests a Skill written by a
practitioner for real work, not one written for the benchmark. The suite
currently holds \ntasks{} tasks in \nscen{} domains, \nmat{} in materials
science and \nbio{} in biology (Table~\ref{tab:tasks}), and is extended on a
rolling basis as further contributions are integrated; the counts in this
draft describe its present state.

For each task we say why the problem matters, what the agent is given and
asked for, what the accompanying Skill is for, and how a result is verified.
In keeping with Section~\ref{sec:skill-conditions}, a Skill is described by
its purpose only. How a Skill achieves that purpose is its contributor's and
is not reproduced here.

\begin{table}[t]
  \centering
  \small
  \caption{Task suite in its present state: \ntasks{} tasks in \nscen{}
  domains.}
  \label{tab:tasks}
  \begin{tabularx}{\linewidth}{@{}>{\raggedright\arraybackslash}p{2.5cm}*{3}{>{\raggedright\arraybackslash}X}@{}}
    \toprule
    Task & Given to the agent & Asked for & How it is verified \\
    \midrule
    \multicolumn{4}{@{}l}{\textit{Materials science} (\nmat)} \\
    \addlinespace[2pt]
    Oxygen tracer diffusion from isotope-exchange SIMS &
      raw $^{16}$O and $^{18}$O line-scan or depth-profile data with
      acquisition metadata; the fitting software &
      the tracer diffusion coefficient $D^{*}$ and surface exchange
      coefficient $k^{*}$, with a review of the surface position, the fit
      residuals and the data quality, and a statement of how far the result
      can be trusted &
      point-by-point residuals of the fit and their RMS; fitted parameters
      and their 95\% confidence intervals against reference values on data
      with known ground truth; whether the data determine $D^{*}$ and $k^{*}$
      separately; checks of parsing and unit handling; diagnostic
      findings against annotated cases, where reporting insufficient evidence
      can be the correct answer \\
    \addlinespace
    Phase identification and unit-cell refinement from powder XRD &
      a powder diffraction pattern and the elements present in the sample;
      access to a crystal-structure database~\citep{grazulis2009cod} and to
      GSAS-II~\citep{toby2013gsas} &
      the crystalline phase and its refined unit-cell parameters from a
      Le~Bail fit~\citep{lebail1988}, with fit statistics and a refinement
      figure &
      fit quality by $R_{\mathrm{wp}}$ (below 10\%) and $\chi^{2}$ (the
      closer to one the better); identified phase against the reference
      phase; refined cell parameters against reference values within
      tolerance; required artifacts present \\
    \addlinespace
    \multicolumn{4}{@{}l}{\textit{Biology} (\nbio)} \\
    \addlinespace[2pt]
    Regulatory target prioritisation &
      a table of candidate enhancer--gene pairs in one cell context with
      published predictive scores, from a public CRISPR perturbation
      study~\citep{fulco2019abc}; experimental outcomes withheld &
      a ranking of candidate regulatory links for each gene, an evidence
      summary that states what is missing, and a proposed follow-up
      experiment &
      the frozen ranking is scored against the withheld outcomes by average
      precision, reported with its coverage; predictions must cover exactly
      the supplied records; the agent runs without access to the outcome
      files \\
    \addlinespace
    Spheroid structure and cell distribution from 3D microscopy &
      a three-dimensional fluorescence microscopy image stack of a cell
      spheroid, with its axes, its voxel spacing in physical units and a
      record of what each channel shows &
      a statement of which analyses the image supports; the envelope of the
      spheroid as a three-dimensional mask and surface, with volume and
      surface area; the positions of the nuclear centres; the density of
      nuclei as a function of distance from the centre and of depth below
      the surface; the distances between neighbouring nuclei; and images for
      reviewing each result &
      envelope against a manually drawn reference (overlap, surface distance,
      volume error); nuclear centres matched one to one with annotated
      centres; required artifacts present; no claim the image cannot support
 \\
    \bottomrule
  \end{tabularx}
\end{table}

\subsection{Materials science}

\subsubsection{SIMS data analysis: oxygen tracer diffusion from isotope exchange}
\label{sec:task-sims}

\paragraph{Why it matters.}
Secondary-ion mass spectrometry (SIMS) answers two questions about an element
inside a solid: how much of it is there, and where it sits. In semiconductor
manufacturing those are the questions asked of every dopant, since a chip
works only if the right amount of the right element has gone to the right
depth, and SIMS depth profiling is how that is checked. The same measurement
serves materials research more broadly. In the task here it follows an oxygen
isotope into an oxide, which gives the rate at which oxygen moves through the
material and across its surface: the tracer diffusion coefficient $D^{*}$ and
the surface exchange coefficient $k^{*}$. In both settings the raw data are
easy to collect and easy to misread.

\paragraph{The task.}
The agent receives the raw $^{16}$O and $^{18}$O data from an
isotope-exchange experiment, as a line scan or a depth profile, together with the acquisition
metadata and the fitting software a researcher would use. It is asked for
$D^{*}$ and $k^{*}$, and for an account of how far those values can be
trusted: whether the data are sound, whether the surface has been placed
correctly, and whether the fit describes the measurement. What makes the task
hard is that a fit can converge and still be wrong. Producing two numbers is
straightforward; knowing when not to believe them is the skill being tested.

\paragraph{The Skill.}
The Skill supports the analysis of SIMS data. It is designed for researchers
who need to know how much of an element is in a material and how it is
distributed, whether that is a dopant in a chip material or an isotope tracer
in an oxide. It helps them turn raw line scans and depth profiles into
quantities they can defend, by checking the quality of the data, reviewing
the fit, and judging how far the result can be trusted.

\paragraph{Verification.}
The result of this task is continuous, and so is its assessment. The fitted
curve and the measured profile are both one-dimensional, so the first
criterion is the residual at every point of the profile: an array with one
entry per depth, which shows not only how large the misfit is but where it
lies. The second summarises that array in one number, the root-mean-square
(RMS) of the residuals, taken after each residual is divided by the
uncertainty of its data point so that a value near one means the fit agrees
with the data to within their noise. The third concerns the reported
uncertainty: the agent must give a 95\% confidence interval for $D^{*}$ and
for $k^{*}$, and on data whose ground truth is known the verifier checks both
how close the fitted values are to the reference values and whether the
intervals contain them. An agent that reports a tight interval around a wrong
value is penalised for it. The fourth looks at the two parameters together.
$D^{*}$ and $k^{*}$ are fitted jointly and their errors are correlated, so
separate intervals can mislead. The verifier therefore also considers the
joint 95\% confidence region of the pair, and the misfit evaluated over the
plane of $(\log D^{*}, \log k^{*})$ around the optimum: a compact minimum
means the data determine both parameters, while a long valley means they
determine only a combination of the two, and an agent that claims both
values with confidence in the second case is wrong even if its fit is good.
A later version may add a map of the residual at each point of the profile
across the same parameter plane.
These criteria are combined into the task's score
in $[0,1]$ (Section~\ref{sec:metrics}). In addition, parsing
and unit handling are checked deterministically, and the agent's diagnostic
findings are compared with annotated cases; reporting that the evidence is
insufficient is, where true, the correct answer.

\subsubsection{XRD refinement: phase identification and unit-cell parameters}
\label{sec:task-xrd}

\paragraph{Why it matters.}
X-ray diffraction reveals the crystal structure of a material: which phases
are present, and the dimensions of their unit cells. It is how a researcher
finds out whether what was made is what was intended. In semiconductor
manufacturing and thin-film growth the question is whether a grown film has
the expected structure; after a synthesis in the laboratory it is simply what
was made. Phase identification and cell refinement are among the most
frequently performed analyses in materials science, and much routine time
goes into them.

\paragraph{The task.}
The agent receives a powder diffraction pattern and the elements known to be
in the sample, with access to a crystal-structure
database~\citep{grazulis2009cod} and to GSAS-II~\citep{toby2013gsas}. It is
asked to identify the crystalline phase and to report its refined unit-cell
parameters from a Le~Bail fit~\citep{lebail1988}, with fit statistics and a
refinement figure. The difficulty lies in judgement as much as in
computation: several candidate structures may fit a pattern about equally
well, and a numerically good fit does not by itself establish the right
structure.

\paragraph{The Skills.}
Two Skills accompany this task, one for each stage, and together they support
the analysis of a material's crystal structure. The first is designed for
researchers who hold a diffraction pattern and know which elements went into
the sample. It helps them decide which phases the measurement actually
supports, for example whether a grown film is the phase that was intended. The second takes
over once the phase is known. It is designed for researchers who need
unit-cell parameters precise enough to report, and it helps them carry the
refinement through to a result and judge whether the fit is good enough to
stand behind.

\paragraph{Verification.}
The quality of the refinement is assessed with the two indices that
crystallographers themselves report~\citep{mccusker1999rietveld,toby2006rfactors}.
The first is the weighted-profile residual $R_{\mathrm{wp}}$, the weighted
difference between the measured and the calculated pattern as a fraction of
the measured pattern; a refinement is counted as acceptable when
$R_{\mathrm{wp}}$ is below 10\%. The second is the goodness of fit $\chi^{2}$,
the squared ratio of $R_{\mathrm{wp}}$ to the value expected from the
counting statistics of the data: the closer $\chi^{2}$ is to one, the higher
the score, since a value of one means the model explains the data to within
their noise. Each index is first turned into a sub-score between 0 and 1,
\begin{equation*}
  s_{R} \;=\; \max\!\Bigl(0,\; 1 - \frac{R_{\mathrm{wp}}}{20\%}\Bigr),
  \qquad
  s_{\chi} \;=\; \min\!\Bigl(\chi^{2},\; \frac{1}{\chi^{2}}\Bigr),
\end{equation*}
so that $s_{R}$ falls linearly from 1 for a perfect fit to 0 at
$R_{\mathrm{wp}} = 20\%$, passing $0.5$ at the 10\% mark, and $s_{\chi}$ is 1
at $\chi^{2} = 1$ and falls off equally whether $\chi^{2}$ is too large (the
model misses structure in the data) or too small (the model fits noise, or
the uncertainties are overestimated). A third sub-score, $s_{\mathrm{cell}}$,
measures the refined cell parameters against the reference values: it is 1
when they agree and falls linearly to 0 at the stated tolerance. The three
are combined by a weighted sum,
\begin{equation*}
  y \;=\; G \,\bigl( w_{R}\, s_{R} + w_{\chi}\, s_{\chi}
        + w_{\mathrm{cell}}\, s_{\mathrm{cell}} \bigr),
  \qquad w_{R} + w_{\chi} + w_{\mathrm{cell}} = 1,
\end{equation*}
with $w_{R} = w_{\chi} = 0.25$ and $w_{\mathrm{cell}} = 0.5$, which gives the
task's score in $[0,1]$ (Section~\ref{sec:metrics}). The factor $G$ is a
gate: it is 1 if the identified phase matches the reference phase and the
required artifacts are present, and 0 otherwise. The gate is there because a
good fit does not by itself show that the structure is
right~\citep{toby2006rfactors}: a pattern refined well against the wrong
phase earns nothing.

\subsection{Biology}

\subsubsection{Regulatory target prioritisation}
\label{sec:task-regulatory}

\paragraph{Why it matters.}
Most of the genetic variation associated with disease lies outside genes, in
the regulatory elements that control when and where genes are switched on.
Working out which elements regulate which gene, in which cell type, is an
early and expensive step in deciding what to test as a drug target. The
evidence is scattered across perturbation experiments and predictive models,
it is incomplete, and each experiment that would fill a gap costs weeks.
Choosing the next experiment well is where the time is won or lost.

\paragraph{The task.}
The agent receives a table of candidate enhancer--gene pairs in one cell
context, with published predictive scores, drawn from a public CRISPR
perturbation study~\citep{fulco2019abc}. The experimental outcomes are
withheld. It is asked to rank the candidate regulatory links for each gene,
to summarise the evidence and state plainly what is missing, and to propose
the follow-up experiment that would be most informative.

\paragraph{The Skill.}
The Skill supports early-stage drug target discovery. It is designed for
research teams that have identified a candidate gene and want to understand
how it may be regulated in a particular cell type. The Skill helps teams
assess the evidence for potential regulatory relationships, identify gaps in
that evidence, and prioritise directions for experimental follow-up.

\paragraph{Verification.}
The agent's ranking is frozen and scored against the withheld outcomes by
average precision, reported together with its coverage. Predictions must
cover exactly the records supplied, and the agent runs without access to the
outcome files.

\subsubsection{Spheroid imaging: structure and cell distribution in three dimensions}
\label{sec:task-spheroid}

\paragraph{Why it matters.}
A spheroid is a ball of cells grown in three dimensions, a few hundred
micrometres across. Unlike cells on a flat dish, the cells of a spheroid sit
at different depths and receive different amounts of oxygen, nutrients and
drug, as they do in a tumour, which is why spheroids are widely used as
tumour models and in drug screening. What a spheroid tells the researcher is
therefore spatial: how large it is, what shape it has, and where its cells
are, from the rim to the core. Microscopes now produce such image stacks by
the plate, far faster than they can be analysed by hand. The analysis is
also easy to get wrong, because light is lost on its way into dense tissue:
the centre of a spheroid looks dark whether or not cells are there.

\paragraph{The task.}
The agent receives a three-dimensional fluorescence microscopy image stack
of a spheroid, for example from a confocal or light-sheet microscope,
together with what is known about its acquisition: the order of its axes,
the voxel spacing in physical units, and which channel shows the nuclei. It
is first asked to say which analyses the image supports. For those that it
does, it is asked for the envelope of the spheroid, as a three-dimensional
mask and a surface with volume and surface area; for the positions of the
nuclear centres; for the density of nuclei as a function of distance from
the centre and of depth below the surface; and for the distances between
neighbouring nuclei. It must also provide images in which each result can be
checked against the original data.

The difficulty is that every one of these quantities can be computed from
any image, good or bad. A stack without calibrated spacing gives no volume in
physical units, and a two-dimensional picture gives no volume at all. A
count of detected nuclei is not a count of cells. A low density in the core
may be biology or may be an imaging artifact, and nuclei that lie close
together are not thereby cells in contact. The agent has to tell the reader
which claims the data support, and declining an analysis that the image
cannot support is a correct answer.

\paragraph{The Skill.}
The Skill analyses microscopy images of spheroids to measure spheroid
structure and cell distribution. It is designed for researchers who image
spheroids in quantity and need numbers they can compare across wells and
conditions. It helps them check that an image stack is fit for analysis,
delineate the spheroid and reconstruct its envelope and surface in three
dimensions, locate the nuclei, and describe how the cells are arranged, both
radially and relative to their neighbours. Its results are offered as
candidates for the researcher to review, with each stage of the analysis
left open to inspection, and it keeps apart what was measured from what
would need independent confirmation.

\paragraph{Verification.}
The envelope is compared with a manually drawn reference by its overlap, the
distance between the two surfaces and the relative error in volume. Nuclear
centres are matched one to one with manually annotated centres within a
fixed physical distance; where only part of a stack is annotated, only that
part is marked. The radial and neighbourhood tables are compared with the
same quantities computed from the reference. The required artifacts must be
present, and a run fails outright if it reports a three-dimensional
measurement from an image that cannot support one, or states a detected
count as a cell count. The reference annotations are kept where the agent
cannot read them.

\subsection{What the tasks have in common}

The tasks come from different fields, but they make the same demands.
Each depends on specialised software or data that a general-purpose coding
environment does not provide. Each rests on procedure that is written down in
the specialist literature and passed on in laboratories, but is thinly
represented in what a model already knows. And in each, the hard part is not
producing an answer but knowing how far to trust it: recognising a fit that
converged for the wrong reason, a candidate that matches by coincidence, a
measurement that reflects the imaging more than the sample, or evidence too
thin to support a conclusion. This is the kind of knowledge a
Skill is meant to carry, which is why these tasks are a demanding test of
whether Skills help.

\bibliographystyle{plainnat}
\bibliography{references}

@article{li2026skillsbench,
  title   = {{SkillsBench}: Benchmarking How Well Agent Skills Work Across Diverse Tasks},
  author  = {Li, Xiangyi and Liu, Yimin and Chen, Wenbo and others},
  journal = {arXiv preprint arXiv:2602.12670},
  year    = {2026}
}

@misc{anthropic2025skills,
  title        = {Equipping Agents for the Real World with {Agent Skills}},
  author       = {{Anthropic}},
  year         = {2025},
  howpublished = {Anthropic Engineering Blog},
  note         = {Accessed 2026-11}
}

@inproceedings{jimenez2024swebench,
  title     = {{SWE-bench}: Can Language Models Resolve Real-World {GitHub} Issues?},
  author    = {Jimenez, Carlos E. and Yang, John and Wettig, Alexander and Yao, Shunyu and Pei, Kexin and Press, Ofir and Narasimhan, Karthik},
  booktitle = {International Conference on Learning Representations (ICLR)},
  year      = {2024}
}

@inproceedings{mialon2024gaia,
  title     = {{GAIA}: A Benchmark for General {AI} Assistants},
  author    = {Mialon, Gr{\'e}goire and Fourrier, Cl{\'e}mentine and Swift, Craig and Wolf, Thomas and LeCun, Yann and Scialom, Thomas},
  booktitle = {International Conference on Learning Representations (ICLR)},
  year      = {2024}
}

@inproceedings{xie2024osworld,
  title     = {{OSWorld}: Benchmarking Multimodal Agents for Open-Ended Tasks in Real Computer Environments},
  author    = {Xie, Tianbao and Zhang, Danyang and Chen, Jixuan and others},
  booktitle = {Advances in Neural Information Processing Systems (NeurIPS)},
  year      = {2024}
}

@article{wang2024voyager,
  title   = {Voyager: An Open-Ended Embodied Agent with Large Language Models},
  author  = {Wang, Guanzhi and Xie, Yuqi and Jiang, Yunfan and Mandlekar, Ajay and Xiao, Chaowei and Zhu, Yuke and Fan, Linxi and Anandkumar, Anima},
  journal = {Transactions on Machine Learning Research},
  year    = {2024}
}

@article{efron1979bootstrap,
  title   = {Bootstrap Methods: Another Look at the Jackknife},
  author  = {Efron, Bradley},
  journal = {The Annals of Statistics},
  volume  = {7},
  number  = {1},
  pages   = {1--26},
  year    = {1979}
}

@article{hake1998interactive,
  title   = {Interactive-Engagement versus Traditional Methods: A Six-Thousand-Student Survey of Mechanics Test Data for Introductory Physics Courses},
  author  = {Hake, Richard R.},
  journal = {American Journal of Physics},
  volume  = {66},
  number  = {1},
  pages   = {64--74},
  year    = {1998}
}

@inproceedings{styles2024workbench,
  title     = {{WorkBench}: A Benchmark Dataset for Agents in a Realistic Workplace Setting},
  author    = {Styles, Olly and Miller, Sam and Cerda-Mardini, Patricio and Guha, Tanaya and Sanchez, Victor and Vidgen, Bertie},
  booktitle = {Conference on Language Modeling (COLM)},
  year      = {2024}
}

@article{yao2024taubench,
  title   = {$\tau$-bench: A Benchmark for Tool-Agent-User Interaction in Real-World Domains},
  author  = {Yao, Shunyu and Shinn, Noah and Razavi, Pedram and Narasimhan, Karthik},
  journal = {arXiv preprint arXiv:2406.12045},
  year    = {2024}
}

@article{barres2025tau2,
  title   = {$\tau^2$-Bench: Evaluating Conversational Agents in a Dual-Control Environment},
  author  = {Barres, Victor and Dong, Honghua and Ray, Soham and Si, Xujie and Narasimhan, Karthik},
  journal = {arXiv preprint arXiv:2506.07982},
  year    = {2025}
}

@misc{chowdhury2024sweverified,
  title        = {Introducing {SWE-bench Verified}},
  author       = {Chowdhury, Neil and others},
  year         = {2024},
  howpublished = {OpenAI},
  note         = {\url{https://openai.com/index/introducing-swe-bench-verified/}}
}

@inproceedings{merrill2026terminalbench,
  title     = {{Terminal-Bench}: Benchmarking Agents on Hard, Realistic Tasks in Command Line Interfaces},
  author    = {Merrill, Mike A. and others},
  booktitle = {International Conference on Learning Representations (ICLR)},
  year      = {2026},
  note      = {arXiv:2601.11868}
}

@inproceedings{patil2025bfcl,
  title     = {The {Berkeley Function Calling Leaderboard} ({BFCL}): From Tool Use to Agentic Evaluation of Large Language Models},
  author    = {Patil, Shishir G. and others},
  booktitle = {Proceedings of the 42nd International Conference on Machine Learning},
  series    = {Proceedings of Machine Learning Research},
  volume    = {267},
  pages     = {48371--48392},
  year      = {2025}
}

@inproceedings{chen2025acebench,
  title     = {{ACEBench}: A Comprehensive Evaluation of {LLM} Tool Usage},
  author    = {Chen, C. and others},
  booktitle = {Findings of the Association for Computational Linguistics: EMNLP 2025},
  pages     = {12970--12998},
  year      = {2025}
}

@article{zhong2025complexfuncbench,
  title   = {{ComplexFuncBench}: Exploring Multi-Step and Constrained Function Calling under Long-Context Scenario},
  author  = {Zhong, Lucen and Du, Zhengxiao and Zhang, Xiaohan and Hu, Haiyi and Tang, Jie},
  journal = {arXiv preprint arXiv:2501.10132},
  year    = {2025}
}

@article{li2025drafterbench,
  title   = {{DrafterBench}: Benchmarking Large Language Models for Tasks Automation in Civil Engineering},
  author  = {Li, Yinsheng and Dong, Zhen and Shao, Yi},
  journal = {arXiv preprint arXiv:2507.11527},
  year    = {2025}
}

@inproceedings{chen2026browsecompplus,
  title     = {{BrowseComp-Plus}: A Fair and Disentangled Evaluation Benchmark for Deep Search Agents},
  author    = {Chen, Z. and others},
  booktitle = {Proceedings of the 64th Annual Meeting of the Association for Computational Linguistics (Volume 1: Long Papers)},
  pages     = {22349--22370},
  year      = {2026}
}

@article{bigeard2025financeagent,
  title   = {{Finance Agent Benchmark}: Benchmarking {LLMs} on Real-world Financial Research Tasks},
  author  = {Bigeard, Antoine and Nashold, Langston and Krishnan, Rayan and Wu, Shirley},
  journal = {arXiv preprint arXiv:2508.00828},
  year    = {2025}
}

@inproceedings{dagan2025plancraft,
  title     = {Plancraft: An Evaluation Dataset for Planning with {LLM} Agents},
  author    = {Dagan, Gautier and Keller, Frank and Lascarides, Alex},
  booktitle = {Conference on Language Modeling (COLM)},
  year      = {2025}
}

@inproceedings{kwa2025horizon,
  title     = {Measuring {AI} Ability to Complete Long Software Tasks},
  author    = {Kwa, Thomas and others},
  booktitle = {Advances in Neural Information Processing Systems},
  volume    = {38},
  year      = {2025}
}

@inproceedings{zou2025art,
  title     = {Security Challenges in {AI} Agent Deployment: Insights from a Large Scale Public Competition},
  author    = {Zou, Andy and others},
  booktitle = {Advances in Neural Information Processing Systems, Datasets and Benchmarks Track},
  volume    = {38},
  year      = {2025}
}

@article{dziemian2026ipi,
  title   = {How Vulnerable Are {AI} Agents to Indirect Prompt Injections? Insights from a Large-Scale Public Competition},
  author  = {Dziemian, M. and others},
  journal = {arXiv preprint arXiv:2603.15714},
  year    = {2026}
}

@inproceedings{yehudai2026survey,
  title     = {A Survey on Evaluation of {LLM}-based Agents},
  author    = {Yehudai, Asaf and Eden, Lilach and Li, Alan and Uziel, Guy and Zhao, Yilun and Bar-Haim, Roy and Cohan, Arman and Shmueli-Scheuer, Michal},
  booktitle = {Findings of the Association for Computational Linguistics: ACL 2026},
  pages     = {26690--26714},
  year      = {2026}
}

@inproceedings{rabanser2026reliability,
  title     = {Towards a Science of {AI} Agent Reliability},
  author    = {Rabanser, Stephan and Kapoor, Sayash and Kirgis, Peter and Liu, Kangheng and Utpala, Saiteja and Narayanan, Arvind},
  booktitle = {Proceedings of the 43rd International Conference on Machine Learning (ICML)},
  year      = {2026}
}

@inproceedings{zhang2025cybench,
  title     = {Cybench: A Framework for Evaluating Cybersecurity Capabilities and Risks of Language Models},
  author    = {Zhang, Andy K. and others},
  booktitle = {International Conference on Learning Representations (ICLR)},
  year      = {2025}
}

@article{zhang2025bountybench,
  title   = {{BountyBench}: Dollar Impact of {AI} Agent Attackers and Defenders on Real-World Cybersecurity Systems},
  author  = {Zhang, Andy K. and others},
  journal = {arXiv preprint arXiv:2505.15216},
  year    = {2025}
}

@inproceedings{wang2026cybergym,
  title     = {{CyberGym}: Evaluating {AI} Agents' Cybersecurity Capabilities with Real-World Vulnerabilities at Scale},
  author    = {Wang, Zhun and others},
  booktitle = {International Conference on Learning Representations (ICLR)},
  year      = {2026},
  note      = {arXiv:2506.02548}
}

@inproceedings{shi2026cybergyme2e,
  title     = {{CyberGym-E2E}: Scalable Real-World Benchmark for {AI} Agents' End-to-End Cybersecurity Capabilities},
  author    = {Shi, Tianneng and others},
  booktitle = {Proceedings of the 43rd International Conference on Machine Learning},
  series    = {Proceedings of Machine Learning Research},
  volume    = {306},
  year      = {2026},
  note      = {arXiv:2606.04460}
}

@article{zhang2026mobilecybench,
  title   = {{MobileCybench}: Evaluating Agent Vulnerability Discovery via Executable Probes},
  author  = {Zhang, Andy K. and others},
  journal = {arXiv preprint arXiv:2609.23980},
  year    = {2026}
}

@article{fulco2019abc,
  title   = {Activity-by-Contact Model of Enhancer--Promoter Regulation from Thousands of {CRISPR} Perturbations},
  author  = {Fulco, Charles P. and Nasser, Joseph and Jones, Thouis R. and others},
  journal = {Nature Genetics},
  volume  = {51},
  number  = {12},
  pages   = {1664--1669},
  year    = {2019}
}

@article{toby2013gsas,
  title   = {{GSAS-II}: The Genesis of a Modern Open-Source All Purpose Crystallography Software Package},
  author  = {Toby, Brian H. and Von Dreele, Robert B.},
  journal = {Journal of Applied Crystallography},
  volume  = {46},
  number  = {2},
  pages   = {544--549},
  year    = {2013}
}

@article{grazulis2009cod,
  title   = {Crystallography Open Database -- An Open-Access Collection of Crystal Structures},
  author  = {Gra{\v{z}}ulis, Saulius and Chateigner, Daniel and Downs, Robert T. and others},
  journal = {Journal of Applied Crystallography},
  volume  = {42},
  number  = {4},
  pages   = {726--729},
  year    = {2009}
}

@article{lebail1988,
  title   = {Ab-initio Structure Determination of {LiSbWO$_6$} by {X}-ray Powder Diffraction},
  author  = {Le Bail, Armel and Duroy, Huguette and Fourquet, Jean-Louis},
  journal = {Materials Research Bulletin},
  volume  = {23},
  number  = {3},
  pages   = {447--452},
  year    = {1988}
}

@article{toby2006rfactors,
  title   = {{R} Factors in {R}ietveld Analysis: How Good Is Good Enough?},
  author  = {Toby, Brian H.},
  journal = {Powder Diffraction},
  volume  = {21},
  number  = {1},
  pages   = {67--70},
  year    = {2006},
  doi     = {10.1154/1.2179804}
}

@article{mccusker1999rietveld,
  title   = {Rietveld Refinement Guidelines},
  author  = {McCusker, L. B. and Von Dreele, R. B. and Cox, D. E. and Lou{\"e}r, D. and Scardi, P.},
  journal = {Journal of Applied Crystallography},
  volume  = {32},
  number  = {1},
  pages   = {36--50},
  year    = {1999},
  doi     = {10.1107/S0021889898009856}
}

@misc{agentskills2025spec,
  title  = {Agent Skills: Specification},
  author = {{Agent Skills}},
  year   = {2025},
  url    = {https://agentskills.io},
  note   = {Accessed 2026-11}
}

@misc{claudecode2026skills,
  title  = {Extend {Claude} with Skills},
  author = {{Anthropic}},
  year   = {2026},
  url    = {https://code.claude.com/docs/en/skills},
  note   = {Claude Code documentation. Accessed 2026-11}
}

@misc{claudecode2026memory,
  title  = {How {Claude} Remembers Your Project},
  author = {{Anthropic}},
  year   = {2026},
  url    = {https://code.claude.com/docs/en/memory},
  note   = {Claude Code documentation. Accessed 2026-11}
}

@misc{codex2026skills,
  title  = {Agent Skills},
  author = {{OpenAI}},
  year   = {2026},
  url    = {https://developers.openai.com/codex/skills},
  note   = {Codex documentation. Accessed 2026-11}
}

@misc{codex2026agentsmd,
  title  = {Custom Instructions with {AGENTS.md}},
  author = {{OpenAI}},
  year   = {2026},
  url    = {https://developers.openai.com/codex/guides/agents-md},
  note   = {Codex documentation. Accessed 2026-11}
}

@misc{opencode2026skills,
  title  = {Agent Skills},
  author = {{OpenCode}},
  year   = {2026},
  url    = {https://opencode.ai/docs/skills},
  note   = {OpenCode documentation. Accessed 2026-11}
}

@misc{opencode2026rules,
  title  = {Rules},
  author = {{OpenCode}},
  year   = {2026},
  url    = {https://opencode.ai/docs/rules},
  note   = {OpenCode documentation. Accessed 2026-11}
}

\end{document}